\documentclass[graybox]{svmult}

\usepackage{type1cm}        
\usepackage{makeidx}         
\usepackage{graphicx}        
\usepackage{multicol}        
\usepackage[bottom]{footmisc}

\usepackage{newtxtext}       %
\usepackage{newtxmath}       

\usepackage{bbding}
\usepackage{multirow}
\usepackage{type1cm}         
\usepackage{amsmath,amssymb}
\usepackage{makeidx}         
\usepackage{graphicx}        
\usepackage{multicol}        
\usepackage[bottom]{footmisc}

\usepackage{newtxtext}       %
\usepackage{newtxmath}       
\usepackage[ruled,vlined,linesnumbered,noresetcount]{algorithm2e}

\makeindex             

\ifodd 0
\newcommand{\rev}[1]{{\color{blue}#1}} 
\newcommand{\com}[1]{\textbf{\color{red}(COMMENT: #1)}} 

\else
\newcommand{\rev}[1]{#1}
\newcommand{\com}[1]{}
\fi

\begin{document}

\title*{Fairness in Learning-Based Sequential Decision Algorithms: A Survey}
\author{Xueru Zhang and Mingyan Liu}
\institute{Xueru Zhang \at University of Michigan, Ann Arbor \email{xueru@umich.edu}
\and Mingyan Liu \at University of Michigan, Ann Arbor \email{mingyan@umich.edu}}
%
%
\maketitle

\abstract{Algorithmic fairness in decision-making has been studied extensively in static settings where one-shot decisions are made on tasks such as classification. However, in practice most decision-making processes are of a sequential nature, where decisions made in the past may have an impact on future data. This is particularly the case when decisions affect the individuals or users generating the data used for future decisions.  In this survey, we review existing literature on the fairness of data-driven sequential decision-making. We will focus on two types of sequential decisions: (1) past decisions have no impact on the underlying user population and thus no impact on future data; (2) past decisions have an impact on the underlying user population and therefore the future data, which can then impact future decisions. 
In each case the impact of various fairness interventions on the underlying population is examined. 
}

\section{Introduction}\label{sec:intro}

Decision-making algorithms that are built from real-world datasets have been widely used in various applications. When these algorithms are used to inform decisions involving human beings (e.g., college admission, criminal justice, resume screening), which are typically done by predicting certain variable of interest from observable features, they may inherit the potential, pre-existing bias in the dataset and exhibit similar discrimination against protected attributes such as race and gender. For example, the COMPAS algorithm used by courts for predicting recidivism in the United States has been shown to be biased against black defendants \cite{dressel2018accuracy}; job searching platform XING ranks less qualified male applicants higher than female applicants who are more qualified \cite{lahoti2019ifair}; a nationwide algorithm used for allocating medical resources in US is biased against black patients \cite{medical}. 

There are various potential causes for such bias. It may have been introduced when data is collected. For instance, if data sampled from a minority group is much smaller in size than that from a majority group, then the model could be more in favor of the majority group due to this representation disparity (e.g., more than a third of data in ImageNet and Open Images, two datasets widely used in machine learning research communities, is US-based \cite{shankar2017no}).  Another example is when the data collection decision itself reflects bias, which then impacts the collected data (e.g., if more police officers are dispatched to places with higher crime rate to begin with, then crimes are more likely to be recorded in these places \cite{ensign2018runaway}). 
Even when the data collection process is unbiased, bias may already exist in the data. Historical prejudice and stereotypes can be preserved in data (e.g., the relationship between "man" and "computer programmers" were found to be similar to that between "woman" and "homemaker" \cite{NIPS2016_6228}).  An interested reader can find more  detailed categorization of bias in the survey \cite{mehrabi2019survey}.  

The problem does not merely stop here. On one hand, decisions made about humans can affect their behavior and reshape the statistics of the underlying population. On the other hand, decision-making algorithms are updated periodically to assure high performance on the targeted populations. This complex interplay between algorithmic decisions and the underlying population can lead to pernicious long term effects by allowing biases to perpetuate and reinforcing pre-existing social injustice. For example, \cite{aneja2019no} shows that incarceration can significantly reduce people's access to finance, which in turn leads to substantial increase in recidivism; this forms a credit-driven crime cycle. Another example is speech recognition: products such as Amazon's Alexa and Google Home are shown to have accent bias with native speakers experiencing much higher quality than non-native speakers \cite{accent_bias}. If this difference in user experience leads to more native speakers using such products while driving away non-native speakers, then over time the data used to train the algorithms may become even more skewed toward native speakers, with fewer and fewer non-native samples. Without intervention, the resulting model may become even more accurate for the former and less for the latter, which then reinforces their respective user experience \cite{pmlr-v80-hashimoto18a}. Similar negative feedback loops have been observed in various settings such as recommendation system \cite{chaney2018algorithmic}, credit market \cite{fuster2018predictably}, and policing prediction \cite{ensign2018runaway}. 
Preventing discrimination and guaranteeing fairness in decision-making is thus both an ethical and a legal imperatives.

To address the fairness issues highlighted above, a first step is to define fairness. Anti-discrimination laws (e.g., Title VII of the Civil Rights Act of 1964) typically assess fairness based on disparate impact and disparate treatment. The former happens when outcomes disproportionately benefit one group while the latter occurs when the decisions rely on sensitive attributes such as gender and race. Similarly, various notions of fairness have been formulated mathematically for decision-making systems and they can be categorized roughly into two classes:
\begin{itemize}
\item \textbf{Individual fairness:} this requires that similar individuals are treated similarly.
\item \textbf{Group fairness:} this requires (approximate) parity of certain statistical measures (e.g., positive classification rate, true positive rate, etc.) across different demographic groups. 
\end{itemize}
In Section \ref{sec:prelim} we present the definitions of a number of commonly used fairness measures. Their suitability for use is often application dependent, and many of them are incompatible with each other \cite{kleinberg2017inherent}. 

To satisfy the requirement of a given definition of fairness, various approaches have been proposed and they generally fall under three categories:
\begin{enumerate}
\item \textbf{Pre-processing:} by changing the original dataset such as removing certain features, reweighing and so on, e.g., \cite{calders2013unbiased,kamiran2012data,zemel2013learning,gordaliza2019obtaining}.
\item \textbf{In-processing:} by modifying the decision-making algorithms such as imposing fairness constraints or changing objective functions, e.g., \cite{berk2017convex,zafar2017fairness,JMLR:v20:18-262,agarwal2018reductions}.
\item \textbf{Post-processing:} by adjusting the output of the algorithms based on sensitive attributes, e.g., \cite{hardt2016equality}.
\end{enumerate}  

While the effectiveness of these approaches have been shown in various domains, most of these studies are done using a static framework where only the immediate impact of the learning algorithm is assessed but not its long-term consequences. Consider an example where a lender decides whether or not to issue a loan based on the applicant's credit score. Decisions satisfying an identical true positive rate (equal opportunity) across different racial groups can make the outcome seem fairer \cite{hardt2016equality}. However, this can potentially result in more loans issued to less qualified applicants in the group whose score distribution skews toward higher default risk. The lower repayment among these individuals causes their future credit scores to drop, which moves the score distribution of that group further toward higher default risk \cite{pmlr-v80-liu18c}. This shows that intervention by imposing seemingly fair decisions in the short term can lead to undesirable results in the long run \cite{xueru1}. As such, it is critical to understand the long-term impacts of fairness interventions on the underlying population when developing and using such decision systems.  

In this survey, we focus on fairness in sequential decision systems. We introduce the framework of sequential decision-making and commonly used fairness notions in Section \ref{sec:prelim}. The literature review is done in two parts.  We first consider sequential settings where decisions do not explicitly impact the underlying population in Section \ref{sec:noeffect}, and then consider sequential settings where decisions and the underlying population interact with each other in Section \ref{sec:interplay}. The impact of fairness interventions is examined in each case.  For consistency of the survey, we may use a set of notations different from the original works.

\section{Preliminaries}\label{sec:prelim}
\subsection{Sequential Decision Algorithms}

The type of decision algorithms surveyed in this paper are essentially classification\rev{/prediction} algorithms used by a decision maker to predict some variable of interest (label) based on a set of observable features. For example, judges predict whether or not a defendant will re-offend based on its criminal records; college admission committee decides whether or not to admit an applicant based on its SAT; lender decides whether or not to issue a loan based on an applicant's credit score. 

To develop such an algorithm, data are collected consisting of both features and labels, from which the best mapping (decision rule) is obtained, which is then used to predict unseen, new data points. Every time a prediction is made, it can either be correct (referred to as a \rev{gain}) or incorrect (referred to as a \rev{loss}). The optimal decision rule without fairness consideration is typically the one that minimizes losses or maximizes gains.

In a sequential framework, data arrive and are observed sequentially and there is feedback on past predictions (loss or gain), and we are generally interested in optimizing the performance of the algorithm over a certain time horizon. 
Such a sequential formulation roughly falls into one of two categories. 

\begin{description}
\item[\textbf{P1:}]  The goal of the algorithm is to learn a near-optimal decision rule 
quickly, noting that at each time step only partial information is available, while minimizing (or maximizing) the total loss (or gain) over the entire horizon.  Furthermore, within the context of fairness, an additional goal is to understand how a fairness requirement impacts such a decision rule. 

\item[\textbf{P2:}]  This is a setting where not only do data arrival sequentially, but decisions made in the past can affect the feature space of the underlying population, thereby changing the nature of future observations.  The goal in this case is to learn an optimal decision rule at each time step and understand the impact it has on the population and 
how fairness requirement further adds to the impact. 
\end{description}


\subsection{Notions of Fairness}\label{sec:fairness}
As mentioned in Section \ref{sec:intro}, different notions of fairness can be generally classified into \textit{individual fairness} and \textit{group fairness}. 

\textbf{Group fairness:} For simplicity of exposition and without loss of generality, we will limit ourselves to the case of two demographic groups $G_a$, $G_b$, distinguished based on some sensitive attribute $Z\in \{a,b\}$ representing group membership (e.g., gender, race). Group fairness typically requires certain statistical measure to be equal across these groups. Mathematically, denote by random variable $Y\in\{0,1\}$ an individual's true label and $\hat{Y}$ its prediction generated from a certain decision rule. Then the following is a list of commonly used group fairness criteria. 
\begin{enumerate}
	
	\item Demographic Parity (\texttt{DP}): it requires the positive prediction rate be equal across different demographic groups, i.e., $\mathbb{P}(\hat{Y} = 1|Z = a) = \mathbb{P}(\hat{Y} = 1|Z = b) $.  
	
	\item Equal of Opportunity (\texttt{EqOpt}): it requires true positive rate (TPR)\footnote{Based on the context, this criterion can also refer to equal false negative rate (FNR), false positive rate
(FPR), or true negative rate (TNR)} be equal across different demographic groups, i.e., $\mathbb{P}(\hat{Y} = 1|Y =1, Z = a) = \mathbb{P}(\hat{Y} = 1|Y=1, Z = b) $. 

\item Equalized Odds (\texttt{EO}): it requires both the false positive rate and true positive rate be equal across different demographic groups, i.e.,  $\mathbb{P}(\hat{Y} = 1|Y =y, Z = a) = \mathbb{P}(\hat{Y} = 1|Y=y, Z = b),\forall y\in\{0,1\}$.

\item Equalized Loss (\texttt{EqLos}): it requires different demographic groups experience the same total prediction error, i.e.,  $\mathbb{P}(\hat{Y} \neq Y|Z = a) = \mathbb{P}(\hat{Y} \neq Y| Z = b)$. 

\end{enumerate}


\textbf{Individual fairness: } Such a criterion targets the individual, rather than group level.  Commonly used examples are as follows. 

\begin{enumerate}
\item Fairness through awareness (\texttt{FA}): this requires that similar individuals be treated similarly.
\item Meritocratic fairness (\texttt{MF}): this requires that less qualified individuals not be favored over more qualified individuals.
\end{enumerate}

The above definitions do not specify how similarity among individuals or qualification of individuals are measured, which can be context dependent. 


\section{(Fair) Sequential Decision When Decisions Do Not Affect Underlying Population}\label{sec:noeffect}
	\begin{table}[h]
\begin{center}
\begin{tabular}{ |c||c|c|c|c|  }
	\hline
	\multirow{2}{ 1.5cm}{ }&\multicolumn{2}{|c|}{Fairness definition}& \multirow{2}{2cm }{  \text{    Data type} }&\multirow{2}{2.2cm }{  Problem type }\\
	\cline{2-3}
	&Group fairness & Individual fairness& & \\
	\hline
	\cite{heidari2018preventing}& & \texttt{FA}&  i.i.d.&\textbf{P1}\\
	\cite{gupta2019individual}   &  & \texttt{FA}& non-i.i.d.&\textbf{P1}\\
	\cite{bechavod2019equal}&  \texttt{EqOpt} & &i.i.d.&\textbf{P1}\\
	\cite{joseph2016fairness} &  &  \texttt{MF} &i.i.d.&\textbf{P1}\\
	 \cite{joseph2018meritocratic}&  & \texttt{MF}&i.i.d.&\textbf{P1}\\
\cite{liu2017calibrated}&  &\texttt{FA} &i.i.d.&\textbf{P1}\\
\cite{ValSinGom18}   &  $ \star$ & &   i.i.d.&\textbf{P1}\\
\cite{blum2018preserving}&   \texttt{EqOpt}, \texttt{EO}, \texttt{EqLos} &    &non-i.i.d.&\textbf{P1}\\
\cite{chen2019fair}&  & $ \star$   &i.i.d.&\textbf{P1}\\
\cite{li2019combinatorial}&   &  $ \star$  &i.i.d.&\textbf{P1}\\
\cite{patil2019achieving}&  &  $ \star$  &i.i.d.&\textbf{P1}\\
 \cite{gillen2018online}&  &  \texttt{FA} &non-i.i.d.&\textbf{P1}\\
	\hline
\end{tabular}
\caption{Summary of related work when decisions do not affect the underlying population. $\star$ represents the use of fairness definitions or interventions not included in Section \ref{sec:fairness}.}
\label{tbl:P1} 
\end{center}
\end{table}

We first focus on a class of sequential decision problems (\textbf{P1}) where the decision at each time step does not explicitly affect the underlying population; a list of these studied are summarized in Table \ref{tbl:P1}.  Most of these works have developed algorithms that can learn a decision rule with sufficient accuracy/performance subject to certain fairness constraint, and the impact of fairness on these sequential decision-making problems is reflected through its (negative) effect on the achievable performance. 

\subsection{Bandits, Regret, and Fair Regret} 

We begin with \cite{heidari2018preventing,gupta2019individual,bechavod2019equal} on online learning problems, where a decision maker at each time $t$ receives data from one individual and makes decision according to some decision rule. It then observes the loss (resp. utility) incurred from that decision. The goal is to learn a decision rule from a set of data collected over $T$ time steps under which (1) the accumulated expected loss (resp. utility) over $T$ steps is upper (resp. lower) bounded; and (2) certain fairness criterion is satisfied. Specifically, \cite{heidari2018preventing,gupta2019individual} focus on individual fairness which ensures that similar individuals (who arrive at different time steps) be treated similarly, by comparing each individual with either all individuals within a time epoch \cite{heidari2018preventing,gupta2019individual} or only those who've arrived in the past \cite{gupta2019individual}.  By contrast, \cite{bechavod2019equal} focuses on group fairness (\texttt{EqOpt}), where at each time the arriving individual belongs to one demographic group and the goal is to ensure different demographic groups in general  receive similar performance over the entire time horizon. Moreover, \cite{bechavod2019equal} considered a partial feedback scenario where the loss (resp. utility) is revealed to the decision maker only when certain decisions are made (e.g., whether an applicant is qualified for a certain job is only known when he/she is hired). \rev{In each of these settings, the impact of fairness constraint on accumulated expected loss/utility is examined and quantified and an algorithm that satisfies both (approximate) fairness and certain loss/utility is developed.}

In some applications, the decision maker at each time makes a selection from multiple choices. For example, hiring employees from multiple demographic groups, selecting candidates from a school for certain competitions, etc.  Specifically, the decision maker at each time receives features of multiple individuals (potentially from different demographic groups) and the corresponding sequential decision problems can be formulated as a multi-armed bandit problem, where each arm represents either one specific individual or one demographic group and choosing an arm represents selecting one individual (from one demographic group). In a classic stochastic bandit problem, there is a set of arms $\mathcal{Z} = \{1, \cdots,K\}$. The decision maker selects an arm $k_t$ at time $t$ from $\mathcal{Z}$ and receives a random reward $r^{k_t}_t$, drawn from a distribution $r^{k}_t\sim \mathbb{P}^k(\cdot;g^k)$ with unknown mean $\mathbb{E}(r_t^{k}) = g^k\in [0,1]$.

Let $h_t = \{(k_s,r_s^{k_s})\}_{s=1}^{t-1}$ represent all history information received by the decision maker up to time $t$.  Then the decision rule $\tau_t$ at $t$ is a probability distribution over all arms.  Denote by $\tau_t(k|h_t)$ the probability of selecting arm $k$ at time $t$ given history $h_t$. The {\em regret} of applying the decision rule $\{\tau_t\}_{t=1}^T$ over $T$ time steps is defined as:
\begin{eqnarray*}
\text{Regret}^T(\{\tau_t\}_t) = \sum_{t=1}^{T}\max_k g^k - \sum_{t=1}^T\mathbb{E}_{k_t\sim \tau_t}\Big[ g^{k_t}\Big] ~. 
\end{eqnarray*}
The goal of a {\em fair} decision maker in this context is to select $\{\tau_t\}_{t=1}^T$ such that the regret over $T$ time steps is minimized, while certain fairness constraint is satisfied.

Joseph et al. in \cite{joseph2016fairness} proposed the use of meritocratic fairness in the above bandit setting as follows. \rev{Consider a multi-armed bandit problem where each arm represents an individual and the decision maker selects one individual at each time.} Let the mean reward $\mathbb{E}(r_t^k)$ represent the average qualification of individuals (e.g., hiring more qualified applicant can bring higher benefit to a company); then it is unfair if the decision maker preferentially chooses an individual less qualified in expectation over another. Formally, the decision maker is defined to be $\delta$-fair over $T$ time steps if with probability $1-\delta$, for all pairs of arms $k,k'\in\mathcal{Z}$ and $\forall t$, the following holds.
\begin{eqnarray}\label{eq:MF}
\tau_t(k|h_t)>\tau_t(k'|h_t)\text{ only if } g^k>g^{k'}~. 
\end{eqnarray}

\cite{joseph2016fairness} developed an algorithm to find optimal decision rules in classic stochastic setting that is $\delta$-fair.  To ensure $\delta$-fairness, for any two arms $k,k'$, they should be selected with equal probability unless $g^k > g^{k'}$. 
Let $u_t^k$, $l_t^k$ be the upper and lower confidence bounds of arm $k$ at time $t$. Then arms $k$ and $k'$ are \textit{linked} if $[l_t^k,u_t^k]\cap [l_t^{k'},u_t^{k'}]\neq \emptyset$; arms $k$ and $k'$ are \textit{chained} if they are in the same component of the transitive closure of the linked relation. The algorithm in \cite{joseph2016fairness} first identifies the arm with the highest upper confidence bound and finds all arms chained to it ($\mathcal{S}_t$). For arms not in $\mathcal{S}_t$, the decision maker has sufficient confidence to claim they are less qualified than others, while for arms in $\mathcal{S}_t$, the decision maker randomly selects one at uniform to ensure fairness. 

\cite{joseph2016fairness} shows that if $\delta < 1/\sqrt{T}$, then the algorithm can achieve  $\text{Regret}^T(\{\tau_t\}_t)= O(\sqrt{K^3T\ln \frac{TK}{\delta}})$. In contrast, without fairness consideration, \rev{the original upper confidence bound (UCB) algorithm proposed by Auer et al. \cite{auer2002finite} achieves regret $\text{Regret}^T(\{\tau_t\}_t)= O(K\log T/\Delta_a)$, where $\Delta_a$ is the difference between the expected rewards of the optimal arm and a sub-optimal arm.} 
The cubic dependence on $K$ (the number of arms) in the former is due to the fact that \rev{any fair decision rule must experience constant per-step regret for $T\gg K^3$ steps on some instances, i.e., the average per-step regret is $\gg 1$ for $T=\Omega(K^3)$. }

The idea of this \textit{chaining strategy} can also be adapted to develop fair algorithms for more general scenarios such as contextual bandit problems \cite{joseph2016fairness} and  bandits with different (or even infinite) number of arms at each time among which multiple arms can be selected \cite{joseph2018meritocratic}. Similar to constraint \eqref{eq:MF}, fairness metrics in these generalized settings are also defined in terms of individual's expected qualification, and stipulate that two similar individuals with the same expected reward be treated similarly, even though their reward distributions can be significantly different. 

In contrast, Liu et al. \cite{liu2017calibrated} proposes smooth fairness based on individuals' reward distributions rather than expected reward, which requires that individuals with similar reward distributions be selected with similar probabilities. Formally, $\forall \epsilon_1,\epsilon_2\geq 0$ and $\forall \delta\in[0,1]$, the decision rule $\tau = \{\tau_t\}_{t=1}^T$ is \textit{$(\epsilon_1,\epsilon_2,\delta)$-smooth fair} w.r.t. a divergence function $D$, if $\forall t$ and for any pair of arms $k,k'$, the following holds with probability at least $1-\delta$: 
\begin{eqnarray}\label{eq:smooth}
D\Big(\text{Ber}(\tau_t(k|h_t))\big|\big|\text{Ber}(\tau_t(k'|h_t))\Big) \leq  \epsilon_1 D\Big(\mathbb{P}^k(\cdot;g^k)\big|\big|\mathbb{P}^{k'}(\cdot;g^{k'})\Big) + \epsilon_2~, 
\end{eqnarray}
where $\text{Ber}(\tau_t(k|h_t))$ denotes a Bernoulli distribution with parameter $\tau_t(k|h_t)$. 

Compared with meritocratic fairness, smooth fairness is weaker in the sense that it allows a worse arm to be selected with higher probability. To quantify such violation, \cite{liu2017calibrated} further proposes a concept of fairness regret, where a violation occurs when the arm with the highest reward realization at a given time is not selected with the highest probability. Based on this idea, the fairness regret of decision rule $\tau_t$ at time $t$ is defined as
\begin{eqnarray*}
R_t^{fair} =  \mathbb{E}\Big[\sum_{k=1}^{K}\max\big(\mathbb{P}^*(k)-\tau_t(k|h_t),0\big)\Big|\{g^k\}_{k=1}^K  \Big]~, 
\end{eqnarray*}
and the cumulative fairness regret is defined as $R^{fair}_{1:T} = \sum_{t=1}^{T}R_t^{fair}$, where $\mathbb{P}^*(k) = \mathbb{P}(k = \underset{k'\in \mathcal{Z}}{\text{argmax}} ~r_t^{k'})$ is the probability that the reward realization of arm $k$ is the highest among all arms. 

Two algorithms were developed in \cite{liu2017calibrated} for special types of bandit problems: (1) Bernoulli bandit, where the reward distributions satisfy $\mathbb{P}^k(\cdot;g^k) = \text{Ber}(g^k)$; and (2) Dueling bandit: $\forall t$ the decision maker selects two arms $k^1_t, k^2_t$ and only observes the outcome $\textbf{1}(r_t^{k^1_t}>r_t^{k^2_t} )$.  These algorithms satisfy smooth fairness w.r.t. total variation distance with low fairness regret.

\rev{In satisfying \texttt{FA} that similar individuals be treated similarly, one challenge is to define the appropriate context-dependent metric to quantify "similarity". Most studies in this space assume such a metric is given. \cite{gillen2018online} proposes to learn such a similarity metric from the decision process itself. Specifically, it considers a linear contextual bandit problem where each arm corresponds to an unknown parameter $\theta\in\mathbb{R}^d$. At each time $t$ the decision maker observes $K$ arbitrarily and possibly adversarially selected contexts $x_t^1,  \cdots, x_t^K\in\mathbb{R}^d$ from $K$ arms, each representing features of an individual. It selects one (say arm $i$) among them according to some decision rule $\tau_t$ and receives reward $r_t^i$ with mean $\mathbb{E}(r^i_t) = \langle x_t^i,\theta \rangle$.  \cite{gillen2018online} focuses on individual fairness that individuals with similar contexts (features) be selected with similar probabilities, i.e., $|\tau_t(k|h_t)-\tau_t(k'|h_t)|\leq D(x_t^k,x_t^{k'})$, $\forall k,k'$, for some unknown metric $D(\cdot,\cdot)$. Similar to \cite{liu2017calibrated}, \cite{gillen2018online} also defines a fairness regret to quantify fairness violation over $T$ time steps. Specifically, let $R^{fair}_{t}(\Delta) = \sum_{i=1}^{K-1}\sum_{j=i+1}^{K}\textbf{1}(|\tau_t(i|h_t)-\tau_t(j|h_t)|> D(x_t^i,x_t^{j})+\Delta )$ be the total number of arm pairs violating $\Delta$-fairness and the total fairness regret over $T$ steps is $R^{fair}_{1:T}(\Delta) = \sum_{t=1}^TR^{fair}_{t}(\Delta)$, where $\Delta$ represents the error tolerance. The goal is to find a decision rule with low fairness regret that is also near-optimal (w.r.t. the best fair decision rule). 
	
However, since $D(\cdot,\cdot)$ is unknown, to achieve the above objective, $D(\cdot,\cdot)$ also needs to be learned. To do so, it assumes that in addition to reward $r_t^i$, the decision maker at each time receives feedback $\{(k,k'):|\tau_t(k|h_t)-\tau_t(k'|h_t)|> D(x_t^k,x_t^{k'}) \}$, i.e., the set of all pairs of individuals for which the decision rule violates the fairness constraint. With such (weak) feedback, a computationally efficient algorithm is developed in \cite{gillen2018online} that for any metric $D(\cdot,\cdot)$ following the form of Mahalanobis distance, i.e., $D(x_1,x_2) = ||Ax_1-Ax_2||_2$ for some matrix $A$, any time horizon $T$ and any $\Delta$, with high probability it (i) obtains regret $\tilde{O}(K^2d^2\log(T)+d\sqrt{T})$ w.r.t. the best fair decision rule; and (ii) violates unknown fairness constraints by more than $\Delta$ on at most $O(K^2d^2\log(d/\Delta))$  steps. }

\rev{Other studies, such as \cite{chen2019fair,li2019combinatorial,patil2019achieving} also use a bandit formulation with fairness consideration, where the fairness constraint requires either each arm be pulled for at least a certain fraction of the total available steps, or the selection rate of each arm be above a threshold. Algorithms that satisfy both (approximate) fairness and low regret are developed in these studies.} 

\subsection{Fair Experts and Expert Opinions} 
In some sequential decision problems, decision maker at each time may follow advice from multiple experts $\mathcal{V} = \{v_k\}_{k=1}^K$ and at each $t$ it selects expert according to a decision rule $\tau_t$ where $\tau_t(k)$ denote the probability of selecting expert $v_k$ at time $t$. Blum et al. \cite{blum2018preserving} considers a  sequential setting where at each time a set of experts $\mathcal{V}$ all make predictions about an individual (possibly based on sensitive attribute $Z_t\in\{a,b\}$). Let $Y_t\in\{0,1\}$ be the individual's true label and expert $v_k$'s prediction be $\hat{Y}_t^{k}$, then the corresponding loss of expert $v_k$ is measured as $l(Y_t,\hat{Y}_t^k)\in[0,1]$. By following decision rule $\tau_t$, the decision maker takes $v_k$'s advice with probability $\tau_t(k)$, and the overall expected loss at time $t$ is given by $\sum_{v_k\in\mathcal{V}}\tau_t(k)l(Y_t,\hat{Y}_t^k)$. The decision maker is assumed to observe $l(Y_t,\hat{Y}_t^k)\in[0,1]$, $\forall v_k\in\mathcal{V}$ and $\forall t$. 

In \cite{blum2018preserving}, each expert in isolation is assumed to satisfy certain fairness criterion $\mathcal{C}\in \{\texttt{EqOpt}, \texttt{EO}, \texttt{EqLos}\}$ over a horizon. Specifically, given a sequence of individuals $\{(y_t,z_t)\}_{t=1}^T$, let $\mathcal{T}_z^y = \{t| z_t = z, y_t = y\}$ be the set of time steps at which corresponding individuals are from $G_z$ and have label $y\in\{0,1\}$, expert $v_k$ satisfies \texttt{EqOpt} if $ \frac{1}{|\mathcal{T}_a^1|} \sum_{t\in \mathcal{T}_a^1} l(y_t,\hat{y}_t^{k}) =\frac{1}{|\mathcal{T}_b^1|} \sum_{t\in \mathcal{T}_b^1} l(y_t,\hat{y}_t^{k}) $ holds. The decision maker following $\tau = \{\tau_t\}$ is said to be $\Delta$-fair w.r.t. \texttt{EqOpt} if the following holds,
$$ \Big| \mathbb{E}\big[  \frac{1}{|\mathcal{T}_a^1|} \sum_{t\in \mathcal{T}_a^1}\sum_{v_k\in\mathcal{V}} \tau_t(k)l(Y_t,\hat{Y}_t^{k})\big] - \mathbb{E}\big[  \frac{1}{|\mathcal{T}_b^1|} \sum_{t\in \mathcal{T}_b^1}\sum_{v_k\in\mathcal{V}} \tau_t(k)l(Y_t,\hat{Y}_t^{k})\big] \Big| \leq \Delta~.$$
Similar formula can be derived for the \texttt{EO} and \texttt{EqLos} criteria. The goal of the decision maker is to find $\Delta$-fair $\tau$ w.r.t. $\mathcal{C}$ from a set of fair experts that all satisfy fairness $\mathcal{C}$ in isolation, and at the same time perform as (approximate) good as the best expert in hindsight. Formally, define $\epsilon$-approximate regret of $\tau$ over $T$ time steps with respect to decision maker $v_i\in\mathcal{V}$ as follows:
\begin{eqnarray}
\text{Regret}^T(\tau,v_i,\epsilon) = \sum_{t=1}^{T} \sum_{v_k\in\mathcal{V}}\tau_t(k) l(y_t,\hat{y}_t^{k}) - (1+\epsilon)  \sum_{t=1}^{T} l(y_t,\hat{y}_t^{i}) ~. 
\end{eqnarray}
Then the goal is to achieve vanishing regret $\mathbb{E}[\text{Regret}^T(\tau,v_i,\epsilon)] = o(T)$, $\forall \epsilon>0$ and $\forall v_i\in\mathcal{V}$. 

When the input is i.i.d., the above setting is trivial because the best expert can be learned in $O(\log |\mathcal{V}|)$ rounds and the decision maker can follow its advice afterwards. Because each expert is fair in isolation, this also guarantees vanishing \rev{discrimination}. 

However, when input is non-i.i.d., achieving such objective is challenging. \cite{blum2018preserving} considers an adversarial setting where both $Z_t$ and $Y_t$ can be adaptively chosen over time according to $\{(Z_s,Y_s,\hat{Y}_s)\}_{s=1}^{t-1}$. It first examines the property of \texttt{EqOpt} and shows that given a set of experts that satisfies \texttt{EqOpt}, it is impossible to find a decision rule $\tau$ with vanishing regret that can also preserve $\Delta$-fairness w.r.t. \texttt{EqOpt}. This negative result holds for both the cases when group identity information $Z_t$ is used in determining $\tau$ (group-aware) and the cases when the group information is not used (group-unaware). Specifically, for both cases, \cite{blum2018preserving} constructs scenarios (about how an adversarial selects $(Z_t,Y_t)$ over time) under which for any $\Delta$ that is smaller than a constant $c<0.5$, $\exists \epsilon>0$ such that for any $\tau$ that satisfies $\mathbb{E}[\text{Regret}^T(\tau,v_i,\epsilon)] = o(T)$, $\forall v_i\in\mathcal{V}$, violates the $\Delta$-fairness w.r.t. \texttt{EqOpt}.  

Since \texttt{EqOpt} is strictly weaker than \texttt{EO}, the above impossibility result in \texttt{EqOpt} naturally generalizes to \texttt{EO}. In contrast, under \texttt{EqLos}, given a set of experts that satisfies \texttt{EqLos} fairness, $\forall \Delta>0$, there exists group-aware $\tau$ that can simultaneously attain $\Delta$-fairness and the vanishing regret. The idea is to run two separate multiplicative weights algorithms for two groups. Because one property of the multiplicative weights algorithm is that it performs no worse than the best expert in hindsight but also no better. Therefore the average performance of each group is approximately equal to the average performance attained by the best expert for that group. Because each expert is \texttt{EqLos} fair, the average performance attained by best experts of two groups are the same. Consequently, both vanishing regret and $\Delta$-fairness are satisfied. This positive result is due to the consistency between performance and fairness measure for \texttt{EqLos}. 
However, such positive result does not generally hold for \texttt{EqLos}. If only one multiplicative algorithm is performed without separating two groups, i.e., run in group-unaware manner, then it can be shown that $\forall \epsilon>0$ and $\forall \Delta>0$, any algorithm satisfying vanishing regret also violates $\Delta$-fairness w.r.t. \texttt{EqLos}.

Valera et al. \cite{ValSinGom18} studied a matching problem in sequential framework, where a set of experts $\mathcal{V}$ need to make predictions about $m$ individuals from two demographic groups over $T$ time steps, where at time step $t$ individual $i$'s decision is made by expert $v_i(t)\in\mathcal{V}$. Different from \cite{blum2018preserving} where experts are all fair (w.r.t. a particular metric) over a horizon and at each time only one expert's advice is followed on one individual, experts in \cite{ValSinGom18} can be biased and at each time predictions from $m$ decision makers are all used and each is assigned to one individual. The algorithms for finding the optimal assignments are developed for cases with and without fairness intervention, \rev{which can improve both the overall accuracy and fairness as compared to random assignment, and fairness is guaranteed even when a significant percentage (e.g., 50\%) of experts are biased against certain groups.}  

\subsection{Fair Policing} 
Ensign et al. \cite{ensign2018runaway} studied a predictive policing problem, where the decision maker at each time decides how to allocate patrol officers to different areas to detect crime based on historical crime incident data. The goal is to send officers to each area in numbers proportional to the true underlying crime rate of that area, i.e., areas with higher crime rate  are allocated more officers. \cite{ensign2018runaway} first characterizes the long-term property of existing predictive policing strategies (e.g., PredPol software), in which \rev{more} officers are sent to areas with the \rev{higher} predicted crime rates and the resulting incident data is fed back into the system. By modeling this problem using $\text{P}\acute{\text{o}}\text{lya}$ urn model, \cite{ensign2018runaway} shows that under such a method, one area can eventually consume all officers, even though the true crime rates may be similar across areas. This is because by allocating more officers to an area, more crimes are likely to be detected in that area; allocating more officers based on more detected crimes is thus not a proper method. To address this issue, effective approaches are proposed in \cite{ensign2018runaway}, e.g., by intentionally normalizing the detected crime rates according to the rates at which police are sent.   

\section{(Fair) Sequential Decision When Decisions Affect Underlying Population}\label{sec:interplay}
\begin{table}
\begin{center}
	\begin{tabular}{ |c||c|c|c|c|  }
		\hline
		\multirow{2}{ 1.5cm}{ }&\multicolumn{2}{|c|}{Fairness notion}&\multirow{2}{2.2cm }{  Problem type }\\
		\cline{2-3}
		&Group fairness & Individual fairness&  \\
		\hline
	\cite{pmlr-v80-liu18c}    &  \texttt{EqOpt}, \texttt{DP} & &\textbf{P2}\\
	\cite{heidari2019long} &  $\star$&    &\textbf{P2}\\
	\cite{kannan2019downstream} &  \texttt{EqOpt}, $\cdots$& &\textbf{P2}\\
\cite{mouzannar2019fair}    & \texttt{DP} & &\textbf{P2}\\
\cite{liu2019disparate} & $\star$ & &\textbf{P2}\\
\cite{hu2018short}  &  \texttt{DP},$\cdots$& &\textbf{P2}\\
\cite{pmlr-v80-hashimoto18a} &$\star$ & &\textbf{P2}\\
		\cite{xueru}& \texttt{EqOpt}, \texttt{DP} & &\textbf{P2}\\
			\cite{jabbari2017fairness}&  & \texttt{MF}&\textbf{P1}\\	
			
			\cite{wen2019fairness}&   \texttt{DP}&&\textbf{P1}\\	
		\hline
	\end{tabular}
\caption{Summary of related work when decisions affect underlying population. $\star$ represents some other fairness notions or interventions that are not introduced in Section \ref{sec:fairness}.}
\label{tbl:P2} 
\end{center}
\end{table}

We next examine a second class of sequential decision problems (\textbf{P2}) where the decisions affect the underlying population; a list of these studied are summarized in Table \ref{tbl:P2}.  We will start with a set of papers that use a two-stage model, followed by a set of papers focusing on finite-horizon and infinite-horizon models. 

\subsection{Two-Stage Models}

To examine the long term impact of fairness intervention on the underlying population, some studies \cite{pmlr-v80-liu18c,heidari2019long,kannan2019downstream} construct two-stage models, whereby the first stage decisions (under certain fairness criterion) are imposed on individuals from two demographic groups $G_a$, $G_b$, which may cause individuals to take certain actions, and the overall impact of this one-step intervention on the entire group is then examined in the second stage. 

Let $\alpha_k$ be the size of $G_k$ as the fraction of the entire population and $\alpha_a + \alpha_b = 1$. \cite{pmlr-v80-liu18c} focuses on a one-dimensional setting where an individual from either group has feature $X\in\mathcal{X}$ with $\mathcal{X} = \{1,2,\cdots,M\}$ and sensitive attribute $Z\in\{a,b\}$ representing his/her group membership. Let $\pi(x|k) = \mathbb{P}(X=x|Z=k), x\in \mathcal{X}$ be $G_k$'s feature distribution and $Y\in\{0,1\}$ the individual's true label.  The decision maker makes predictions on individuals using the decision rule $\tau(x,k) = \mathbb{P}(\hat{Y} = 1|X=x,Z=k)$ and receives expected utility $u(x)$ for making a positive prediction $\hat{Y}=1$ of an individual with feature $x$ (e.g., average profit of a lender by issuing a loan to applicants whose credit score is 760). The expected utility of the decision maker under $\tau$ is given by:
\begin{align*}
U(\tau) = \sum_{k\in\{a,b\}} \alpha_k \sum_{x\in\mathcal{X}} u(x) \tau(x,k)\pi(x|k)~. 
\end{align*}
Define the \textit{selection rate} of $G_k$ under a decision rule as $\gamma(k) = \mathbb{P}(\hat{Y}=1|Z=k)= \sum_{x\in\mathcal{X}} \tau(x,k)\pi(x|k)$.  Then given feature distributions, the relationship between $\gamma(k)$ and $\tau(\cdot,k)$ can be described by an invertible mapping $g(\cdot)$ so that $\gamma(k) = g(\tau(\cdot,k);\pi(\cdot,k))$ and $\tau(\cdot,k) = g^{-1}(\gamma(k);\pi(\cdot,k))$.

In \cite{pmlr-v80-liu18c}, decision rules for $G_a$, $G_b$ are selected such that $U(\tau)$ is maximized under fairness constraints defined as follows: 
\begin{itemize}
	\item \texttt{Simple}: it requires the same decision rule be used by $G_a$, $G_b$, i.e., $\tau(\cdot,a) = \tau(\cdot,b)$.
	\item Demographic Parity (\texttt{DP}): it requires the selection rates of $G_a$, $G_b$ are equalized, i.e., $\gamma(a)  = \gamma(b)$.
	\item Equal of Opportunity (\texttt{EqOpt}): it requires the true positive rate (TPR) of $G_a$, $G_b$ are equalized, i.e., $\mathbb{P}(\hat{Y}=1|Y=1,Z=a)=\mathbb{P}(\hat{Y}=1|Y=1,Z=b)$ .
\end{itemize}

Once an individual with feature $X=x$ is predicted as positive ($\hat{Y}=1$) in the first stage, its feature may be affected; denote the average of such change as $\Delta(x)$. For example, consider a lending scenario where a lender decides whether or not to issue loans to applicants based on their credit scores. Among applicants who are issued loans, those with the higher (resp. lower) credit score are more likely to repay (resp. default); as a result, the credit scores may increase for applicants who can repay the loans ($\Delta(x)>0$) but decrease for those who default ($\Delta(x)<0$). Consequently, the feature distribution of the entire group can be skewed. Let the impact of a decision rule $\tau(x,k)$ on $G_k$ be captured by the average change of $X$ in $G_k$, defined as $\Delta \mu(\tau,k)= \sum_{x\in\mathcal{X}} \tau(x,z)\pi(x|k)\Delta(x)$.
It can be shown that $\Delta \mu(\tau,k)$ is a concave function in the selection rate $\gamma(k)$. 

Let the optimal fair decision rule that maximizes $U(\tau)$ under fairness criterion $\mathcal{C}\in \{\texttt{Simple}, \texttt{DP}, \texttt{EqOpt}\}$ be noted as $\tau^{\mathcal{C}}$, and the corresponding selection rate be noted as $\gamma^{\mathcal{C}}$. Let group labels $a,b$ be assigned such that $G_b$ is the disadvantaged group in the sense that $\gamma^{\texttt{Simple}}(a)>\gamma^{\texttt{Simple}}(b)$. 
Given $\Delta \mu(\tau,b)$, a decision rule $\tau$ causes 
\begin{itemize} 
\item \textit{active harm} to $G_b$ if $\Delta \mu(\tau,b)<0$; 
\item \textit{relative harm} if $\Delta \mu(\tau,b)<\Delta \mu(\tau^{\texttt{Simple}},b)$; \item \textit{relative improvement} if $\Delta \mu(\tau,b)>\Delta \mu(\tau^{\texttt{Simple}},b)$. 
\end{itemize} 

Due to the one-to-one mapping between the decision rule and the selection rate, the notation $\Delta \mu(\tau,k) = \Delta \mu(g^{-1}(\gamma(k);\pi(\cdot,k)),k)$ in the following is simplified as $\Delta \tilde{\mu}(\gamma(k),k)$. Let $\gamma_0(b)$ be the \textit{harmful} threshold for $G_b$ such that $\Delta \tilde{\mu}(\gamma_0(b),b)=0$; let $\gamma^*(b)$ be the \textit{max-improvement} threshold such that $\gamma^*(b) = {\text{argmax}_{\gamma}}~\Delta \tilde{\mu}(\gamma,b)$; let $\bar{\gamma}(b)$ be the \textit{complementary} threshold such that $\Delta \tilde{\mu}(\bar{\gamma}(b),b) = \Delta \tilde{\mu}(\gamma^{\texttt{Simple}}(b),b)$ and $\gamma^{\texttt{Simple}}(b) < \bar{\gamma}(b)$.

The goal of \cite{pmlr-v80-liu18c} is to understand the impact of imposing \texttt{DP} or \texttt{EqOpt} fairness constraint on $\Delta \mu(\tau,k)$, whether these fairness interventions can really benefit the disadvantaged group $G_b$ as compared to the \texttt{Simple} decision rule.

\cite{pmlr-v80-liu18c} first examined the impact of \texttt{Simple} decision rule, and showed that if $u(x)>0 \Longrightarrow \Delta(x) >0 $, then \texttt{Simple} threshold does not cause active harm, i.e., $\Delta \mu(\tau^{Simple},b)\geq 0$. In lending example, the condition $u(x)>0 \Longrightarrow \Delta(x) >0 $ means that the lender takes a greater risk by issuing a loan to an applicant than the applicant does by applying.

For \texttt{DP} and \texttt{EqOpt} fairness, \cite{pmlr-v80-liu18c} showed that both could cause relative improvement, relative harm and active harm, under different conditions. We summarize these results below, for $\mathcal{C} \in \{\texttt{DP}, \texttt{EqOpt}\}$,
\begin{enumerate}
\item Under certain conditions, there exists $\alpha_0<\alpha_1<1$ such that $\forall \alpha_b\in[\alpha_0,\alpha_1]$, $\tau^{\mathcal{C}}$ causes relatively improvement, i.e., ${\gamma}^{\texttt{Simple}}(b)< \gamma^{\mathcal{C}}(b) <\bar{\gamma}(b)$.
\item Under certain conditions, positive predictions can be over-assigned to $G_b$ for satisfying $\mathcal{C}$. There exists $\alpha_0$ such that $\forall \alpha_b\in [0,\alpha_0]$, $\tau^{\mathcal{C}}$ causes relatively harm or active harm, i.e., $\gamma^{\mathcal{C}}> \bar{\gamma}(b)$ or $\gamma^{\mathcal{C}}> \gamma_0(b)$. 
\end{enumerate}

These results show that although it seems fair to impose \texttt{DP} and \texttt{EqOpt} constraints on decisions (e.g., by issuing more loans to the disadvantaged group), it may have unintended consequences and harm the disadvantaged group (e.g., features of disadvantaged group may deteriorate after being selected). 

\cite{pmlr-v80-liu18c} makes further comparisons between \texttt{DP} and \texttt{EqOpt} fairness. Generally speaking, \texttt{DP} and \texttt{EqOpt} cannot be compared in terms of $\Delta \mu(\tau,b)$.  Because there exist both settings when \texttt{DP} causes harm while \texttt{EqOpt} causes improvement, and settings when \texttt{EqOpt} causes harm while \texttt{DP} causes improvement. However, for some special cases when $\pi(\cdot|a)$ and $\pi(\cdot|b)$ satisfy a specific condition, there exists $\alpha_0,\alpha_1$ such that $\forall \alpha_b\in[\alpha_0,\alpha_1]$, \texttt{DP} causes active harm while \texttt{EqOpt} causes improvement. 
Moreover, if under \texttt{Simple} decision rule, ${\gamma}^{\texttt{Simple}}(a) > {\gamma}^{\texttt{Simple}}(b)$ and $\mathbb{P}(\hat{Y}=1|Y=1,Z=b)>\mathbb{P}(\hat{Y}=1|Y=1,Z=a)$ hold, then $\gamma^{\texttt{EqOpt}}(b)<{\gamma}^{\texttt{Simple}}(b)<\gamma^{\texttt{DP}}(b)$ can be satisfied, i.e., \texttt{EqOpt} can cause relative harm by selecting less than \texttt{Simple} rule.

An interested reader is referred to \cite{pmlr-v80-liu18c} for details of the specific conditions mentioned above. It shows that temporal modeling and a good understanding of how individuals react to decisions are necessary to accurately evaluate the impact of different fairness criteria on the population. 

\subsubsection{Effort-based Fairness} 
Essentially, the issues of unfairness described in the preceding section may come from the fact that different demographic groups have different feature distributions, leading to different treatments. However, this difference in feature distributions is not necessarily because one group is inherently inferior to another; rather, it may be the result of the fact that advantaged group can achieve better features/outcomes with less effort. For example, if changing one's school type from public to private can improve one's SAT score, then such change would require much higher effort for the low-income population. From this point of view, Heidari et al. \cite{heidari2019long} proposes an effort-based notion of fairness, which measures unfairness as the disparity in the average effort individuals from each group have to exert to obtain a desirable outcome. 

Consider a decision maker who makes a prediction about an individual using decision rule $h(\cdot)$ based on its $d$-dimensional feature vector $X\in\mathcal{X}$.  Let $Y\in\mathcal{Y}$ be the individual's true label, $Z\in\{a,b\}$ its sensitive attribute, and $\hat{Y} = h(X)$ the predicted label. Define a benefit function $w(h(X),Y)\in \mathbb{R}$ that quantifies the benefit received by an individual with feature $X$ and label $Y$ if he/she is predicted as $h(X)$.

For an individual from $G_k$ who changes his/her data from $(x,y)$ to $(x',y')$, the total effort it needs to take is measured as $E_k\big((x,y),(x',y')\big) = \frac{1}{d}\sum_{i=1}^d e_{k}^i(x_i,x_i')$,
where $x = (x_1,\cdots,x_d)$, $x' = (x'_1,\cdots,x'_d)$ and $e_{k}^i(x_i,x_i')$ denotes the effort needed for an individual from $G_k$ to change its $i$th feature from $x_i$ to $x'_i$. 
Accordingly, the change in the individual's benefit by making such an effort is $\Delta w\big((x,y),(x',y')\big) = w(h(x),y) - w(h(x'),y')$, 
and the total utility received by an individual from $G_k$ in changing his/her data is
\begin{eqnarray*}
U_k\big((x,y),(x',y')\big) = \Delta w\big((x,y),(x',y')\big) - E_k\big((x,y),(x',y')\big)~. 
\end{eqnarray*}

Define $\widehat{U}_k = \mathbb{E}\big[\max_{(x',y')\in \mathcal{X}\times \mathcal{Y}} U_k\big((x,y),(x',y')\big)  |Z=k\big]$ as the expected highest utility $G_k$ can possibly reach by exerting effort.  \cite{heidari2019long} suggests the use of the disparity between $\widehat{U}_a$ and $\widehat{U}_b$ as a measure of group unfairness. 


The microscopic impact of decisions on each individual can be modeled using the above unfairness measure. Intuitively, if individuals can observe the behaviors of others similar to them, then they would have more incentive to imitate behaviors of those (social models) who receive higher benefit, as long as in doing so individuals  receive positive utility.

Let $\mathcal{D}_k$ be the training dataset representing samples of population in $G_k$.  Then
$(x^*,y^*) =  \underset{(x',y')\in\mathcal{D}_k }{\text{argmax}}U_k\big((x,y),(x',y')\big)$ can be regarded as a social model's profile that an individual $(x,y)$ from $G_k$ aims to achieve, as long as $U_k\big((x,y),(x^*,y^*)\big)>0$. Given the change of each individual in $\mathcal{D}_k$, a new dataset $\mathcal{D}_k'$ in the next time step can be constructed accordingly.

Given $\mathcal{D}_k$, $\mathcal{D}_k'$, the datasets before and after imposing decisions according to $h(\cdot)$, the macroscopic impact of decisions on the overall underlying population can be quantified. \cite{heidari2019long} adopts the concept of segregation from sociology to measure the degree to which multiple groups are separate from each other. Specifically, the segregation of $\mathcal{D}_k$ and $\mathcal{D}_k'$ are compared from three perspectives: {\em Evenness, Clustering} and {\em Centralization}. The details of each can be found in \cite{heidari2019long}; here we only introduce {\em Centralization} as an example: this is measured as the proportion of individuals from a minority group whose prediction $h(X)$ is above the average. The impact of decisions on the entire group is examined empirically by comparing {\em Evenness, Clustering} and {\em Centralization} of $\mathcal{D}_k$ and $\mathcal{D}_k'$.

\cite{heidari2019long} first trained various models $h(\cdot)$ such as neural network, linear regressor and decision tree over a real-world dataset without imposing a fairness constraint. It shows that individuals by imitating social model's data profile can either increase or decrease the segregation 
of the overall population, and different models may shift the segregation toward different directions. Next, \cite{heidari2019long} examined the impact of imposing fairness constraint on a linear regression model. Specifically, the fairness constraint requires each group's average utility be above the same threshold, a higher threshold indicating a stronger fairness requirement. Empirical results show that segregation under different levels of fairness can change in completely different directions (decrease or increase), and impacts on {\em Evenness, Centralization} and {\em Clustering} are also different. 

Indeed, fairness intervention affects segregation in two competing ways. If more desirable outcomes are assigned to a disadvantaged group intentionally, then on one hand individuals from the disadvantaged group may have less motivation to change their features, on the other hand, the same individuals may serve as social models, which in turn can incentivize others from the same disadvantaged group to change their features. Both impacts are at play simultaneously and which one is dominant depends on the specific context. This paper highlights the fact that modifying decision algorithm is not the only way to address segregation and unfairness issues; imposing mechanisms before individuals enter the decision system may be another effective way, e.g., by decreasing the costs for individuals from the disadvantaged group to change their features. 

\subsubsection{A Two-Stage Model in College Admissions} 

Kannan et al. \cite{kannan2019downstream} studied a two-stage model in the case of college admissions and hiring.  In the first stage, students from two demographic groups are admitted to a college based on their entrance exam scores; in the second stage an employer chooses to hire students from those who were admitted to the college based on their college grades. Specifically, let $Z\in\{a,b\}$ denote a student's group membership and $Y\sim \mathcal{N}(\mu_k,\sigma^2_k), k\in\{a,b\}$ his/her qualification drawn from a group-specific Gaussian distribution. Let variable $X = Y+ \nu$ be the student's entrance exam score with independent noise $\nu\sim \mathcal{N}(0,1),\forall k\in\{a,b\}$.

Denote by $\hat{Y}^c\in\{0,1\}$ the college's admissions decision about a student.  Let $\tau^c(x,k) = \mathbb{P}(\hat{Y}^c=1|X=x,Z=k)\in[0,1]$ be the admissions rule representing the probability a student from $G_k$ with score $x$ gets admitted, which is monotone non-decreasing in $x$ for $k\in\{a,b\}$. Consider a threshold decision rule of the following form:
\begin{eqnarray}\label{eq:college}
\tau^c(x,k) = \begin{cases}
1, \text{ if }x \geq \theta_k\\
0, \text{ if } x< \theta_k
\end{cases}
\end{eqnarray}

For a student who is admitted, he/she receives a grade $G = Y+\mu$ with $\mu\sim \mathcal{N}(0,\sigma^2_c)$, $\forall k\in\{a,b\}$, where the variance $\sigma^2_c>0$ is determined by some grading rule. Specifically, $\sigma^2_c \rightarrow \infty$ can be regarded as a case where students' grades are not revealed to the employer, whereas $\sigma^2_c \rightarrow 0$ represents a case where the employer has perfect knowledge of the students' qualifications.  The employer decides whether or not to hire a student based on his/her grade. Let $c\in [c^-,c^+]$ be the cost for the employer for hiring a student, which can either be known or unknown to the college. Then a student from $G_k$ with grade $g$ gets hired if the employer can achieve a non-negative expected utility, i.e., $\mathbb{E}[Y|G = g,\hat{Y}^c=1,Z=k] \geq c$.

The goal of \cite{kannan2019downstream} is to understand what admission rules and grading rules should be adopted by the college in the first stage so that the following fairness goals may be attained in the second stage:
\begin{itemize}
\item Equal of Opportunity (\texttt{EqOpt}): it requires the probability of a student being hired by the employer conditional on the qualification $Y$ is independent of group membership $Z$.
\item Irrelevance of Group Membership (\texttt{IGM}): it requires the employer's hiring decision, conditional on $\hat{Y}^c$ and $G$, should be independent of group membership, i.e., $\forall g\in\mathbb{R}$, $\mathbb{E}[Y|G = g,\hat{Y}^c=1,Z=a] \geq c \Longleftrightarrow \mathbb{E}[Y|G = g,\hat{Y}^c=1,Z=b] \geq c$.
\item \text{Strong Irrelevance of Group Membership (\texttt{sIGM}):} it requires the employer's posterior about students' qualifications, conditional on $\hat{Y}^c$ and $G$, should be independent of group membership, i.e., $\forall g\in\mathbb{R}$ and $\forall y\in\mathbb{R}$,
$\mathbb{P}[Y=y|G = g,\hat{Y}^c=1,Z=a] = \mathbb{P}[Y=y|G = g,\hat{Y}^c=1,Z=b]$.
\end{itemize}

Below we present two simple scenarios found in \cite{kannan2019downstream}, in which both \texttt{EqOpt} and \texttt{IGM} can be satisfied in the second phase under some admission rules.  
\begin{enumerate}
\item Noiseless entrance exam score, i.e., $X=Y$.

In this scenario, the admission decision is determined by the student's qualification $Y$ completely. \cite{kannan2019downstream} shows that as long as the threshold in the admission decision rule is set as $\theta_k = c^+, \forall k\in\{a,b\}$ in Eqn. \eqref{eq:college}, then $\forall [c^-,c^+]\subset \mathbb{R}$ and with any grading rule, both \texttt{EqOpt} and \texttt{IGM} can be satisfied. 

\item No grade is revealed to the employer, i.e., $\sigma^2_c\rightarrow\infty$.

In this case, as long as the threshold in the admission decision rule is set as $\theta_a = \theta_b = \theta$ for some sufficiently large $\theta$ (e.g., highly selective MBA programs) in Eqn. \eqref{eq:college}, then $\forall [c^-,c^+]\subset \mathbb{R}$, both \texttt{EqOpt} and \texttt{IGM} can be satisfied. 
\end{enumerate}

\cite{kannan2019downstream} also studied more general scenarios when noises $\mu$ and $\nu$ are both of finite variance, i.e., noisy entrance exam scores and when colleges report informative grades to the employer.
When employer's hiring cost $c$ is known to the college,  $\forall c\in\mathbb{R}$, there always exist two thresholds $\theta_a^*$, $\theta_b^*$ and a grade $g^*$ for college, under which $\mathbb{E}[Y|G = g^*,X\geq\theta_a^*,Z=a] = \mathbb{E}[Y|G = g^*,X\geq\theta_b^*,Z=b]=c$ always holds, i.e., \texttt{IGM} can always be satisfied. 

However, if we consider the employer's posterior distributions on students' qualification, as long as two groups have different prior distributions, for any two thresholds $\theta_a,\theta_b$ in the admission rule, there always exists $y$ such that $\mathbb{P}[Y=y|G = g,X\geq\theta_a,Z=a] \neq \mathbb{P}[Y=y|G = g,X\geq\theta_b,Z=b]$, i.e., satisfying \texttt{sIGM} is impossible.

Moreover, suppose prior distributions of two groups' qualifications are Gaussian distributed with different mean but the same variance, then $\forall c$, there exists no threshold decision rule $\tau^c$ that can satisfy both \texttt{EqOpt} and \texttt{IGM} simultaneously.  For \texttt{EqOpt} under some fixed hiring cost $c$, in cases when grading rule has variance $\sigma^2_c \neq 1$, there is no threshold decision rule $\tau^c$ such that \texttt{EqOpt} can be satisfied. For cases when $\sigma^2_c = 1$, \texttt{EqOpt} can be satisfied only if the admission rule and grading rule can satisfy $\mathbb{E}[Y|G = \theta_b,X\geq\theta_a,Z=a] = \mathbb{E}[Y|G = \theta_a,X\geq\theta_b,Z=b] = c.$
Such condition is generally impossible to hold. It concludes that  \texttt{EqOpt} is generally is impossible to achieve. 

If employer's hiring cost $c$ is uncertain that college only knows the interval $[c^-,c^+]$, when two groups have different priors, \cite{kannan2019downstream} shows that $\forall c\in [c^-,c^+]$, neither  \texttt{EqOpt} nor \texttt{IGM} can be satisfied  even in isolation under a threshold admission rule. 

The above results show that even with a simple model studied in \cite{kannan2019downstream}, many common and natural fairness goals are impossible to achieve in general.  Such negative results are likely to hold true in more complex models that capture more realistic aspects of the problem. 


\subsection{Long-Term Impacts on the Underlying Population}

Decisions made about humans affect their actions.  
Bias in decisions can induce certain behavior, which is then
captured in the dataset used to develop decision algorithms in the future. The work \cite{pmlr-v80-liu18c,heidari2019long,kannan2019downstream} introduced in the previous section studied such one-step impact of decisions on the population. However, when newly developed algorithms are then used to make decisions about humans in the future, those humans will be affected and biases in the datasets generated by humans can perpetuate.  
 This closed feedback loop becomes self-reinforcing and can lead to highly undesirable outcomes over time. In this section, we focus on the long-term impacts of decisions on population groups.  The goal is to understand what happens to the underlying population when decisions and people interact with each other and what interventions are effective in sustaining equality in the long run.

\subsubsection{Effects of Decisions on the Evolution of Features}

One reason why decisions are made in favor of one group is that the favored group is believed to bring more benefit to the decision maker. For example, a lender issues more loans to a group believed to be more likely to repay, a company hires more from a group perceived to be more qualified, and so on. In other words, disparate treatment received by different groups is due to the disparity in their (perceived) abilities to produce good outcomes (qualifications). From this perspective, the ultimate social equality is attained when different demographic groups possess the same abilities/qualifications. In this section, we present studies reported in \cite{mouzannar2019fair,liu2019disparate,hu2018short} to understand how qualifications of different groups evolve over time under various fairness interventions, and under what conditions social equality may be attained. 

Let $G_a$, $G_b$ be two demographic groups, $\alpha_k$ the size of $G_k$ as a fraction of the entire population and assumed constant, and $\alpha_a + \alpha_b = 1$. Each individual has feature $X$, sensitive attribute $Z\in\{a,b\}$, and label $Y\in\{0,1\}$ representing his/her qualification or the ability to produce certain good outcome. Define the \textit{qualification profile} of $G_k$ at time $t$ as the probability distribution $\pi_t(y|k) = \mathbb{P}_t(Y = y|Z = k)$, $y\in\{0,1\}$.  Changes in feature $X$ induced by decisions are captured by change in the qualification profile. 

Using the definition of qualification profiles of two groups, social equality can be defined formally as equalized qualification profiles, i.e., 
\begin{align}\label{eq:equal}
\lim_{t\rightarrow \infty} |\pi_t(1|a) - \pi_t(1|b)| = 0. 
\end{align}

\cite{mouzannar2019fair,liu2019disparate} assume that the qualification profiles at each time are known to the decision maker, who makes prediction about each individual according to a decision rule 
$\tau_t(y,k) =\mathbb{P}_t(\hat{Y} = 1|Y = y,Z = k)$ and receives utility $u(y)$ for making positive prediction $\hat{Y} = 1$, where $u(0)\leq 0$ and $u(1)\geq 0$ correspond to the loss and benefit, respectively. Define the \textit{selection rate} of $G_k$ under a decision rule at time $t$ as $ \gamma_t(k) = \mathbb{P}_t(\hat{Y}=1|Z = k) = \sum_{y\in\{0,1\}} \gamma_t(y|k)=  \sum_{y\in\{0,1\}} \mathbb{P}_t(\hat{Y}=1|Y=y,Z = k)\mathbb{P}_t(Y=y|Z = k)$. Then the expected utility of the decision maker at $t$ is: 
\begin{align}
U_t(\tau_t) &= \sum_{k\in\{a,b\}} \alpha_k \sum_{y\in \{0,1\}} u(y)\mathbb{P}_t(\hat{Y}=1|Y=y,Z = k)\mathbb{P}_t(Y=y|Z = k) \nonumber\\
& = \sum_{k\in\{a,b\}} \alpha_k \sum_{y\in \{0,1\}} u(y) \tau_t(y,k)\pi_t(y|k)~.
\label{eq:utility}
\end{align}
\rev{Upon receiving a decision, a qualified individual can either remain qualified or become unqualified, and an unqualified individual can either become qualified or remain unqualified for the next time step.  In \cite{mouzannar2019fair}, the evolution of a group's qualification profile is modeled as a dynamical system as follows:}
\begin{align}\label{eq:dynamic1}
\pi_{t+1}(1|k) = \pi_t(1|k)\nu\big(\gamma_t(0|k), \gamma_t(1|k)\big) +  \pi_t(0|k)\mu\big(\gamma_t(0|k), \gamma_t(1|k)\big)
\end{align}
where $\nu(\cdot,\cdot):[0,1]\times[0,1]\rightarrow[0,1]$ represents the retention rate of subgroup who are qualified $(Y=1)$ in time $t$ that are still qualified in $t+1$, while $\mu(\cdot,\cdot):[0,1]\times[0,1]\rightarrow[0,1]$ represents the improvement rate of subgroup who are unqualified ($Y=0$) at time $t$ but make progress to be qualified ($Y=1$) at time $t+1$. \rev{Due to the mapping between the decision rule and the selection rate, the impact of decisions on individuals' future qualifications are captured by the impact of selection rates on the overall qualification profiles via some general functions $\nu(\cdot,\cdot)$ and $\mu(\cdot,\cdot)$ in model \eqref{eq:dynamic1}.}

The goal of the decision maker is to find a decision rule $\tau_t$ with or without fairness consideration, so as to maximize $U_t(\tau_t)$. 
It examines what happens to the qualification profiles of two groups when these decisions are applied at each time, and under what conditions social equality is attained under these decisions.   

Without fairness considerations, the corresponding optimal decision at time $t$ for $G_k$, $k\in\{a,b\}$, is given by\footnote{\rev{Note that such an ideal decision rule assumes the knowledge of $y$, which is not actually observable. In this sense this decision rule, which has 0 error, is not practically feasible. Our understanding is that the goal in \cite{mouzannar2019fair} is to analyze what happens in such an ideal scenario when applying the perfect decision.  }}:
\begin{eqnarray}\label{eq:optUN}
\tau_t^{*}(y,k) = \underset{\tau_t}{\text{argmax}}~ U_t(\tau_t) = \begin{cases}
0, \text{  if  } y = 0\\
1, \text{  if  } y = 1
\end{cases}~. 
\end{eqnarray}
Using this decision rule the selection rate $\gamma_t(k) = \pi_t(1|k)$.  Since the decision rules for the two groups are not constrained by each other, the dynamics \eqref{eq:dynamic1} can be simplified as follows: $\forall k\in \{a,b\}$,
\begin{align}
\pi_{t+1}(1|k) = \Phi(\pi_t(1|k)) \text{ with } \Phi(\pi) = \pi \nu(0,\pi) + (1-\pi) \mu (0,\pi)~. 
\end{align}

Social equality can be attained for any starting profiles $\pi_0(1|a), \pi_0(1|b)$ in this unconstrained case if and only if the system $\pi_{t+1} = \Phi(\pi_t)$ has a unique globally attracting equilibrium point $\pi^*$ and a sufficient condition is given in \cite{mouzannar2019fair}. 



\cite{mouzannar2019fair} also studied impact of fairness intervention on dynamics. It focuses on the notion of demographic parity (\texttt{DP}), which requires the selection rates of two groups to be equal, i.e., $\gamma_t(a) = \gamma_t(b), \forall t$. 
Depending on group proportions $\alpha_a,\alpha_b$ and utilities $u(1)$, $u(0)$, there are two possibilities for the fair optimal decision rule $\tau_t^{*}$. If group labels $a,b$ are assigned such that $G_a$ is the advantaged group, i.e., $\pi_t(1|a)\geq \pi_t(1|b)$, then we have:
\begin{align}
\text{if } \alpha_au(1)+ \alpha_b u(0) \leq 0: & \tau^{*}_t(0,a) = \tau^{*}_t(0,b)  = 0 \label{eq:under1}\\ \text{(under-selected)~~~~~~}
& \tau^{*}_t(1,a) = \frac{\pi_t(1|b)}{\pi_t(1|a)}; \tau^{*}_t(1,b)  = 1 \label{eq:under2}\\
\text{if } \alpha_au(1)+ \alpha_b u(0) \geq 0: &  \tau^{*}_t(0,a) = 0; \tau^{*}_t(0,b)  = \frac{\pi_t(1|a) -\pi_t(1|b)}{1 - \pi_t(1|b)}\label{eq:over1} \\
\text{(over-selected)~~~~~~~}& \tau^{*}_t(1,a) = \tau^{*}_t(1,b)  = 1 \label{eq:over2}
\end{align} 

To guarantee equalized selection rates, Eqn. \eqref{eq:under1}\eqref{eq:under2} show the case where $G_a$ is under-selected, while Eqn. \eqref{eq:over1}\eqref{eq:over2} show the case where $G_b$ is over-selected. Dynamics \eqref{eq:dynamic1} can then be expressed as follows:
\begin{align*}
\pi_{t+1}(1|a) = \Phi_a(\pi_t(1|a),\pi_t(1|b)) \\
\pi_{t+1}(1|b) = \Phi_b(\pi_t(1|a),\pi_t(1|b)) 
\end{align*}
Similar to the unconstrained case, sufficient conditions for reaching social equality in these case can also be derived and are given in \cite{mouzannar2019fair}. 

By comparing these sufficient conditions, \cite{mouzannar2019fair} shows that unconstrained optimal decision rules may reach social equality on its own in some cases. However, if \texttt{DP} fair decisions are used instead in these special cases, then the equality may be violated. Specifically, if disadvantaged group $G_b$ is over selected, social equality may or may not be attained by using \texttt{DP} fair decisions. Moreover, for settings where equality can be attained under both types of decisions, \cite{mouzannar2019fair} further shows that \texttt{DP} fair decisions may lead to the higher total utility as well as the more qualified population in the long run. In contrast, if advantaged group $G_a$ is under selected, social equality will definitely be attained by using \texttt{DP} fair decisions. However, imposing this additional fairness constraint may decrease the decision maker's utility and the population's overall qualification level.

Liu et al. \cite{liu2019disparate} also studied a similar problem on the evolution of qualification profiles of different demographic groups. In their setting, decisions applied to each group can incentivize each individual to rationally invest in his/her qualifications, as long as the expected reward received from the decision maker's prediction outweighs the investment cost.   
 
Formally, the impact of decisions on the underlying population is captured by \textit{individual's best-response}. Let random variable $c_k$ be the cost incurred by an individual from $G_k$ in order to obtain $Y = 1$ (be qualified).  Let cumulative distribution function (CDF) of $c_k$ be denoted as $\mathbb{F}_k(\cdot)$. For any individual, regardless of the group membership $Z$ and actual qualification $Y$, he/she receives a reward $w>0$ only if he/she is predicted as positive (qualified) $\hat{Y}=1$. Therefore, \rev{an individual from $G_k$ at $t$ acquires qualification $Y=1$ if and only if the resulting utility of investing outweighs  the utility of not investing, i.e., }

\begin{eqnarray}
&& \underbrace{w \mathbb{P}_t(\hat{Y}=1|Y = 1,Z = k) - c_k}_{\text{utility if investing}} \rev{-} \underbrace{w \mathbb{P}_t(\hat{Y}=1|Y = 0,Z = k)}_{\text{utility if not investing}} \nonumber \\
&=& w(\tau_t(1,k) - \tau_t(0,k)) - c_k \rev{ > 0}~. 
\label{eqn:util_investment} 
\end{eqnarray}

\rev{Note that the qualification status $Y$ of each individual depends completely on whether he/she invests: given decision rule $\tau_t$, individuals become qualified as long as Eqn. \eqref{eqn:util_investment} is satisfied. The overall qualification profile of $G_k$ is the probability of individuals being qualified, i.e., $\mathbb{P}(Y=1|Z=k)$, or equivalently, the probability of investment cost being sufficiently small (Eqn. \eqref{eqn:util_investment}). Therefore, the update of qualification profile of $G_k$ at $t+1$  can be captured by the CDF of cost variable $c_k$ according to the following:}
\begin{align}\label{eq:qual}
\pi_{t+1}(1|k) = \mathbb{P}(c_k < w(\tau_t(1,k) - \tau_t(0,k)) ) = \mathbb{F}_k (w(\tau_t(1,k) - \tau_t(0,k))  )
\end{align}

\rev{Consider the decision rule that maximizes the decision maker's utility as given in Eqn. \eqref{eq:utility} at each time, i.e., $\tau_t(y,k) = \underset{\tau}{\text{argmax}} ~U_t(\tau)$.  Then the ideal (though infeasible) decision is the same as in Eqn. \eqref{eq:optUN} and is given by the following\footnote{\rev{In \cite{liu2019disparate} the assumption that such a perfect decision rule with 0 error is feasible is formally stated as "realizability".}}, $\forall k\in\{a,b\}$,}
\begin{eqnarray}\label{eq:share}
\tau_t(y,k) =\underset{\tau}{\text{argmax}} ~U_t(\tau) =  \begin{cases}
0, \text{ if } y = 0\\
1, \text{ if } y = 1
\end{cases}
\end{eqnarray}

Given initial qualification profiles $\pi_0(1|a)$ and $\pi_0(1|b)$, $\pi_t(1|k)$ can be captured by a dynamic system $\pi_{t+1}(1|k) = \Phi(\pi_t(1|k))$ for some $\Phi(\cdot)$. We first present the results in \cite{liu2019disparate} under the assumption that CDF of cost variables for two groups are the same, i.e., $\mathbb{F}_a(\cdot) = \mathbb{F}_b(\cdot) = \mathbb{F}(\cdot)$. 

If the perfect decision rule shown in Eqn. \eqref{eq:share} is feasible, then this dynamic system has a unique non-zero equilibrium $\pi^*$ and the corresponding qualification profile $\pi^*(1|k) = \lim_{t\rightarrow \infty}\Phi^t(\pi_0(1|k)) = \mathbb{F}(w)$ is also the optimal for $G_k$.\footnote{$\Phi^t$ is a  $t$-fold composition of $\Phi$.} However, \rev{since this ideal decision is not generally feasible in practice, the evolution of equilibria for more realistic cases} is further examined in \cite{liu2019disparate}.
Let prediction $\hat{Y}$ be calculated from features $X$ via a mapping $h(\cdot): \mathcal{X}\rightarrow\{0,1\}$. \cite{liu2019disparate} focused on two special cases: (1) uniformly distributed $X$; (2) spherical multivariate Gaussian distributed $X$.  For both cases, every group $G_k$ in isolation can be perfectly predicted by some mapping $h_k(\cdot)$ but when both groups are combined, such perfect mapping does not exist.
\cite{liu2019disparate} shows that for both cases, under certain conditions, decision maker by applying $h_k(\cdot)$ to both groups at each time can result in a stable equilibrium at which $\pi^*(1|k) = \mathbb{F}(w) > \pi^*(1|\{a,b\}\setminus k)$, i.e., the qualification profile of $G_k$ is optimal, decision is always in favor of $G_k$ and social equality is violated. Although there exists a unique decision rule $\hat{h}(\cdot)$ for both cases, following which at each time can result in an equilibrium satisfying $\pi^*(1|a) = \pi^*(1|b)$ (social equality), such equilibrium is unfortunately shown to be unstable.

Both cases show that as long as the initial decision rule is not $\hat{h}(\cdot)$, equilibria of the dynamic system can be in favor of one group and biased against the other, social equality cannot be attained. Above results hold under the case when CDF of cost variables for two groups are the same, i.e., $\mathbb{F}_a(\cdot) = \mathbb{F}_b(\cdot)$. If remove this assumption and let $G_b$ be disadvantaged in the sense that its investment cost is sufficiently higher than $G_a$, then \cite{liu2019disparate} shows that there is no stable equilibrium that is in favor of $G_b$ and no equilibrium can result in social equality. This conclusion, although is negative, suggests an effective intervention that can potentially improve the qualification profile of disadvantaged group at the equilibrium: by subsidizing the cost of investment for disadvantaged group. 

Another effective intervention proposed in \cite{liu2019disparate} is by decoupling the decision rules by group, i.e., each group is predicted by its own group-specific decision rule  instead of sharing the same decision rule for all groups. In this case, different from \eqref{eq:share}, at each time $t$ decision maker chooses two decision rules for two groups,
\begin{eqnarray*}
\forall k\in\{a,b\}:~~\tau_t(y,k)= \underset{\tau}{\text{argmax}}~\sum_{y\in\{0,1\}}u(y)\tau(y,k)\pi_t(y|k)
\end{eqnarray*} 
and qualification profile of $G_k$ at $t+1$ is updated in the same way as \eqref{eq:qual}. Under this new dynamic system, \cite{liu2019disparate} shows that $\forall k\in\{a,b\}$, if there exists a perfect decision rule for $G_k$ such that $\tau_t(1,k) = 1$ and $\tau_t(0,k)=0$, then the resulting unique equilibrium $\pi^*$ is stable and satisfy $\pi^*(1|k) =\mathbb{F}(w)$, i.e., both groups have the optimal qualification profiles. If there is no perfect decision rule for both groups, i.e., $\max_{\tau}\sum_{k\in\{a,b\}} \alpha_k (\tau(1,k)-\tau(0,k)) < 1$, then we can still guarantee that at lease one group's qualification profile can be strictly improved at an equilibrium as compared to the case when both groups use the same decision rule.


\subsubsection{Fairness Intervention on Labor Market} 

Hu and Chen \cite{hu2018short} studied the impact of fairness intervention on labor market. In their setting, each individual from either group is a worker. All workers pass through a sequence of two markets: a temporary labor market (TLM) and a permanent labor market (PLM). 
Inside the labor market, workers who get hired by an employer will produce an outcome that is either "good" or "bad". $\forall k\in\{a,b\}$, how well can group $G_k$ perform in general at time $t$ is measured by the group's \textit{reputation}, defined as the proportion of all workers in $G_k$ (including those who are not employed) who can produce "good" outcomes in the labor market over the time interval $ [t-t_0,t]$, noted as $\pi_t^k$. 
Within this context, social equality introduced in Eqn. \eqref{eq:equal} earlier is re-defined: it is attained if the group reputation is equalized, i.e., 
$$ \lim_{t\rightarrow\infty} |\pi_t^a-\pi_t^b| = 0~. $$
\cite{hu2018short} shows that social equality can be attained by imposing short-term fairness intervention in the TLM. Below we take a closer look at this dual labor market model. 

In order to compete for a certain job in the future, worker $i$ from $G_k$ at time $t$ may choose to make education investment $\eta_i\geq 0 $ based on expected wage $w_t$ of the job and its personal cost in the investment $c_{\pi^k_t}(\mu_i,\eta_i)$. The cost depends on two factors: 
\begin{itemize}
\item Worker's ability $\mu_i$: it is an intrinsic attribute of workers with CDF $\mathbb{F}_{\mu}(\cdot)$, which is identical to both groups.

\item Reputation of the group ($\pi^k_t$) that worker $i$ belongs to: workers from a group with better reputation face better cost conditions.
\end{itemize}

Let variable $\rho\in\{Q,U\}$ denote a worker's qualification status and the probability of a worker being qualified for the job is $\gamma(\eta_i)\in[0,1]$ where $\gamma(\cdot)$ is a monotonic increasing function. 
Whether or not a worker can be hired in the TLM is determined by the worker's investment and his/her group membership via a mapping $\tau_{TLM}(\eta_i,k)\in \{0,1\}$. If $\tau_{TLM}(\eta_i,k) = 1$, then worker $i$ that is hired in the TLM is eligible to enter the PLM\footnote{\rev{$\tau_{TLM}=1$ only ensures a worker's eligibility to be hired in the PLM (a necessary condition);  whether the worker is indeed hired in the PLM is determined by the hiring strategy in the PLM.}}. 
Specifically, worker $i$ keeps the same job in the TLM until a Poisson process selects him/her to enter the PLM. Upon entering the PLM, at each time he/she cycles through jobs.  

In order to be hired in the PLM, workers build their own personal reputation $\Gamma^{s}$ by consistently exerting efforts $E$ and producing outcomes $O$ in labor markets. 
 Specifically, workers can exert either high ($H$) or low ($L$) effort with cost $e_{\rho}(\mu_i)$ or 0, and produce either good ($G$) or bad ($B$) outcome. Denote $p_H$ as the probability a worker producing good outcome with high effort and $p_{\rho}$ as the probability a worker producing good outcome with low effort and qualification status $\rho$. 
Worker $i$'s personal reputation $\Gamma_i^s\in[0,1]$ is the proportion that he/she produces good outcomes during the recent length-$s$ history in the labor market, which determines whether or not he/she can be hired at each time in the PLM via a mapping $\tau_{PLM}(\Gamma^s_i)\in\{0,1\}$.

Group $G_k$'s reputation $\pi_{t'}^k$ at time $t'$, which determines the worker's cost in education before entering the TLM, will also be updated based on the outcomes produced by all workers from $G_k$ during time lag $[t'-t_0,t']$. Moreover, the expected wage of the job $w_t$ that determines workers' investments before entering the TLM is also updated in a Poisson manner based on $g_{t'}$, the proportion of workers that are hired in the labor market producing good outcomes in their jobs at $t'<t$.
The above form a feedback loop between the labor market and workers. 

 \cite{hu2018short} studied the long-term impact of imposing fairness constraints in determining $\tau_{TLM}$. Specifically, it compares hiring strategies in the TLM under three constraints:
 \begin{itemize}
\item Demographic Parity (\texttt{DP}): among workers hired in the TLM, a $\alpha_k$ fraction of them are from $G_k$.
\item \texttt{Simple}: both groups are subject to the same hiring strategy, i.e., $\tau_{TLM}(\cdot,a) = \tau_{TLM}(\cdot,b) $.
\item Statistical Discrimination (\texttt{SD}): this is a hiring strategy based on the firm's belief of worker qualifications, e.g., $\mathbb{P}(\rho = Q|Z = k,\eta)= \frac{p_Q(\eta)\rho^p_k}{p_Q(\eta)\rho^p_k + p_U(\eta)(1-\rho^p_k)} $, where $\rho^{p}_k$ denotes the prior of $G_k$' capabilities, and $p_Q(\eta), p_U(\eta)$ denote the probabilities of a qualified and unqualified worker investing $\eta$, respectively. 

 \end{itemize}

\cite{hu2018short} analyzed the optimal hiring strategies of firms in the TLM and PLM, as well as the workers' optimal effort/investment strategy; it also examined the group reputation $(\pi_t^a,\pi_t^b)$ over time when \texttt{DP} fairness intervention is imposed in the TLM. 
They show that there exists a unique stable equilibrium and $T$ such that $\pi_t^a = \pi_t^b , \forall t>T$, i.e., short-term fairness intervention in the TLM can result in two groups gradually approaching the same reputation level and achieving social equality. 
%
Without fairness intervention, workers from the group with better reputation are more likely to invest in education (which is cheaper), enter the PLM and produce good outcomes, which further improves their group reputation. With the \texttt{DP} constraint, the hiring thresholds take into account the differences in costs of investment, and the fractions of workers from two groups that enter PLM are maintained at $\alpha_a, \alpha_b$. 
As a result, the proportions of workers producing good outcomes do not diverge and social equality can be reached.

In contrast, under either the \texttt{Simple} or \texttt{SD} hiring strategy in the TLM, the two groups will not be proportionally represented in the labor market according to $\alpha_a,\alpha_b$ as they have different costs in investment. 
Their group reputations will diverge eventually and cannot reach social equality. 



\subsubsection{Effects of Decisions on Group Representation}

Decision algorithms developed from multiple demographic groups can inherit representation disparity that may exist in the data: the algorithm may be less favorable to groups contributing less to the training process; this in turn can degrade population retention in these groups over time, and exacerbate representation disparity in the long run. Hashimoto et al. \cite{pmlr-v80-hashimoto18a} are among the first to show that the (unconstrained) empirical risk minimization (ERM) formulation, which is widely used in training machine learning models, can amplify group representation disparity over time. 

Consider two demographic groups $G_a$, $G_b$.  An individual from either group has  feature $X\in\mathcal{X}$ and label $Y\in\mathcal{Y}$. Let $f_a(x,y)$ and $f_b(x,y)$ be the joint distributions of $(X,Y)$ for individuals in $G_a$ and $G_b$, respectively. At each time $t$, a decision maker receives data $\mathcal{D}_t$ from a set of individuals. Specifically, $\forall k\in\{a,b\}$, let $N_k(t)$ be the expected number of individuals in $\mathcal{D}_t$ that are from $G_k$ and $\alpha_k(t) = \frac{N_k(t)}{N_a(t)+N_b(t)}$ is how much $G_k$ is represented in the data.  Then the overall feature distribution of the entire population at $t$ is given by $f_t(x,y) = \alpha_a(t) f_a(x,y) + \alpha_b(t) f_b(x,y)$. Denote $\alpha(t) = [\alpha_a(t);\alpha_b(t)]$. 

Let $h_{\theta}:\mathcal{X}\rightarrow\mathcal{Y}$ be a decision rule for predicting label from features, which is parameterized by some parameter $\theta\in \mathbb{R}^d$. Let $l(h_{\theta}(X),Y)$ be the prediction loss incurred by predicting $(X,Y)$ using $h_{\theta}(\cdot)$ where $l(\cdot, \cdot)$ is the loss function measuring the discrepancy between predictions and true labels. The goal of the decision maker at time $t$ is to find a $\theta(t)$ for both groups such that the overall prediction loss is minimized:
\begin{align}\label{eq:prob}
\theta(t)=\theta(\alpha(t)) = \underset{\theta}{\text{argmin}} ~L(\theta) =  \mathbb{E}_{(X,Y)\sim f_t(x,y)} [ l(h_{\theta}(X),Y)]~. 
\end{align}


Individuals after receiving their predictions may choose to either leave the decision system or stay. For those who experience low accuracy, they have a higher probability of leaving the system. As a result, the impact of decisions on the overall group representation can be captured by a discrete-time user retention model:
\begin{align}
N_k(t+1) &= \Phi(N_k(t)) = N_k(t)\cdot \nu(L_k(\theta(\alpha(t)))) + \beta_k \label{eq:dynamic}\\
\alpha_k(t+1) &= \frac{N_k(t+1)}{N_a(t+1)+N_b(t+1)}\nonumber
\end{align}
where $L_k(\theta(\alpha(t))) = \mathbb{E}_{(X,Y)\sim f_{k}(x,y)} [l(h_{\theta(t)}(X),Y)]$ is the expected loss experienced by $G_k$ from decision $\theta(t)$, retention rate $\nu(\cdot)\in [0,1]$ represents the probability of an individual who was in system at $t$ remaining in the system at $t+1$, and $\beta_k$ is the number of new users from $G_k$. 

Under the systems given in Eqn. \eqref{eq:prob}\eqref{eq:dynamic}, \cite{pmlr-v80-hashimoto18a} first finds the condition under which a fixed point of the system 
is unstable;  
 the representation disparity under such unstable systems will be amplified over time.

To prevent one group from diminishing, or, to ensure $\alpha_k(t)>\alpha_{\min}, \forall t$, for some $\alpha_{\min}$, instead of minimizing the overall prediction loss, \cite{pmlr-v80-hashimoto18a} suggests bounding the worst-case group loss $L_{\max}(\theta(\alpha(t))) = \max\{L_a(\theta(\alpha(t))),L_b(\theta(\alpha(t)))\}$, $\forall t$.  This can be challenging as the true sensitive attribute $Z$ of each data point is unknown to the decision maker. To address this, a distributionally
robust optimization (DRO) is formulated in \cite{pmlr-v80-hashimoto18a}. Instead of controlling $L_{\max}(\theta(\alpha(t)))$ directly, it controls an upper bound on it. Specifically, it considers the worst-case loss among all perturbed distributions $\tilde{f}_r(x,y)$ that are within a chi-squared ball $\mathcal{B}(f(x,y),r)$ around real distribution $f(x,y)$. Let $\mathcal{B}(f(x,y),r) = \{\tilde{f}_r(x,y) | D_{\chi^2}(f||\tilde{f}_r)\leq r\}$ where $D_{\chi^2}(f||\tilde{f}_r)$ is $\chi^2$-divergence between distributions $f(x,y)$ and $\tilde{f}_r(x,y)$, then $\forall \theta$ and $f_t(x,y)$, loss experienced  by $G_k$ can be upper bounded by: $$L_{dro}(\theta;r_k) = \underset{\tilde{f}_r(x,y)\in \mathcal{B}(f_t(x,y),r_k)}{\sup}\mathbb{E}_{(X,Y)\sim \tilde{f}_r(x,y)}[l(h_{\theta}(X),Y)]\geq \mathbb{E}_{(X,Y)\sim {f}_k(x,y)}[l(h_{\theta}(X),Y)] $$
with robustness radius $r_k = (1/\alpha_k(t)-1)^2$. Consequently, $L_{\max}(\theta(\alpha(t)))$ can be controlled by choosing 
\begin{eqnarray}\label{eq:dro}
\theta(\alpha(t)) = \underset{\theta}{\text{argmin}} ~L_{dro}(\theta;r_{\max})
\end{eqnarray}
 with $r_{\max} = (1/\min\{\alpha_a(t),\alpha_b(t) \}-1)^2$.
 
Suppose $\forall k\in\{a,b\}$, the initial states satisfy $\alpha_k(1)>\alpha_{\min}$. Using the above method, \cite{pmlr-v80-hashimoto18a} shows that $\alpha_k(t)>\alpha_{\min},\forall t$, can be guaranteed for the entire horizon under the following condition:
 $$L_{dro}(\theta(\alpha(t));r_{\max})\leq \nu^{-1}\left( 1- \frac{(1-\nu_{\max})\beta_k}{\alpha_{\min}(\beta_a+\beta_b) }\right)~,$$ 
where $\nu_{\max} = \max \{\nu(L_a(\theta(t))),\nu(L_b(\theta(t)))\}$. While the above condition is hard to verify in practice, experiments in \cite{pmlr-v80-hashimoto18a} show that the decisions selected according to the DRO formulation \eqref{eq:dro} result in  stronger stability of group representation than that selected by ERM formulation \eqref{eq:prob}.
 
 \cite{pmlr-v80-hashimoto18a} shows that the group representation disparity can worsen over time when no fairness is imposed when making a decision. In contrast, Zhang et al. \cite{xueru} show that it can worsen even when fairness criteria are imposed.  
They consider a similar sequential framework where at each time $t$ two (potentially different) decision rules $h_{\theta_a(t)}(\cdot), h_{\theta_b(t)}(\cdot)$ are applied to $G_a$, $G_b$ and parameters $\theta_a(t)$, $\theta_b(t)$ are selected to optimize an objective, subject to certain fairness criterion $\mathcal{C}$: 
 \begin{align}
 \underset{(\theta_a,\theta_b )}{\text{  argmin}}~&\pmb{O}_t(\theta_a,\theta_b;\alpha_a(t),\alpha_b(t)) = \alpha_a(t) O_{a,t}(\theta_a) + \alpha_b(t) O_{b,t}(\theta_b)\label{eq:objO}\\
\text{s.t. }&\Gamma_{\mathcal{C},t}(\theta_a,\theta_b) = 0\nonumber
 \end{align}
Note that the overall objective at time $t$ consists of sub-objectives from two groups weighted by their group proportions at $t$, and empirical risk minimization \eqref{eq:prob} studied in \cite{pmlr-v80-hashimoto18a} is a special case of \eqref{eq:objO}, with $\theta_a = \theta_b$ and $O_{k,t}(
 \theta_k) = L_k(\theta)$ being $G_k$'s empirical loss $\forall t$. 
 Similar to \cite{pmlr-v80-hashimoto18a}, group representation is affected by decisions according to a user retention model and are updated over time,
 \begin{align}
 N_k(t+1) &= N_k(t)\cdot \pi_{k,t}(\theta_k(t)) + \beta_k \label{eq:dynamicour}\\
 \alpha_k(t+1) &= \frac{N_k(t+1)}{N_a(t+1)+N_b(t+1)}\nonumber ~. 
 \end{align}
 As compared to \eqref{eq:dynamic}, the retention rate $\pi_{k,t}(\theta_k(t))$ of $G_k$ can be any function that depends on the decision, which means the analysis and conclusions obtained in \cite{xueru} are not limited to applications where user retention is driven by model accuracy (e.g., speech recognition, medical diagnosis); instead they are more generally applicable (e.g., in lending/hiring, user retention is more likely to be driven by positive classification rate rather than the expected loss.)
 
The goal of \cite{xueru} is to characterize long-term property of group representation disparity $\frac{\alpha_a(t)}{\alpha_b(t)}$, and understand what is the impact of imposing various fairness constraints in this process.  
It turns out that even with fairness intervention, group representation disparity can still change monotonically and one group may diminish over time from the system. Specifically, given a sequence of one-shot problems $\{{\pmb{O}_t}(\theta_a,\theta_b;{\alpha}_a(t),{\alpha}_b(t)) \}_{t=1}^{T}$, if $\forall t$, ${\pmb{O}_t}$ is defined over the same sub-objectives ${O}_{a}(\theta_a)$, ${O}_{b}(\theta_b)$ with different group proportions $(\alpha_a(t),\alpha_b(t))$, and the dynamics satisfy $\pi_{k,t}(\theta_k) = h_k(O_k(\theta_k))$ for some decreasing function $h_k(\cdot)$, i.e., user departure is driven by the value of sub-objective function, then the group representation disparity $\frac{\alpha_a(t)}{\alpha_b(t)}$ changes monotonically over time and the discrepancy between $\pi_{a,t}(\theta_a(t))$ and $\pi_{b,t}(\theta_b(t))$ increases over time. Intuitively, whenever one group's proportion (e.g., $\alpha_a(t)$) starts to increase, the decision maker in minimizing the overall objective would select a decision pair such that $O_a(\theta_a(t))$ decreases. Consequently, $G_a$'s retention as determined by $h_a(O_a(\theta_a(t)))$ increases, i.e., $G_a$'s proportion increases further and representation disparity worsens.  

This condition that leads to exacerbating representation disparity can be easily satisfied under commonly used objectives (e.g., minimizing overall expected loss), common fairness constraints (e.g., \texttt{EqOpt}, \texttt{DP}, etc.), and various dynamics (e.g., user participation driven by model accuracy or intra-group disparity); an interested reader is referred to \cite{xueru} for more details. 
It highlights the fact that common fairness interventions fail to preserve representation parity. This is ultimately because what are being equalized by those fairness criteria often do not match what drives user retention; thus applying seemingly fair decisions may worsen the situation.  A main takeaway is that fairness must be defined with a good understanding of the underlying user retention model, which can be challenging in practice as we typically have only incomplete/imperfect information. However, if user dynamics model is available, \cite{xueru} presents the following method for finding a proper fairness criterion that mitigates representation disparity.

Consider a general dynamics model $N_k(t+1) = \Phi\big(N_k(t),\{\pi^m_k(\theta_k(t))\}_{m=1}^M,\beta_k\big)$, $\forall k\in\{a,b\}$, where user departures and arrivals are driven by $M$ different factors $\{\pi^m_k(\theta_k(t))\}_{m=1}^M$ (e.g., accuracy, false positive rate, positive rate, etc.). Let $\Theta$ be the set of all possible decisions, if there exists a pair of decisions $(\theta_a,\theta_b)\in\Theta\times\Theta$ under which dynamics have stable fixed points, then a set $\mathcal{C}$ of decisions $(\theta_a,\theta_b)$ that can sustain group  representation can be found via an optimization problem:
\begin{align*}
\mathcal{C} = \underset{(\theta_a,\theta_b)}{\text{argmin}}& ~ \Big|\frac{\widetilde{N}_a}{\widetilde{N}_b} - \frac{\beta_a}{\beta_b} \Big|\\
 \text{s.t. }& \widetilde{N}_k = \Phi\big(\widetilde{N}_k,\{\pi^m_k(\theta_k)\}_{m=1}^M,\beta_k\big)\in \mathbb{R}_{+}, \theta_k \in \Theta, \forall k\in\{a,b\}~. 
\end{align*}
The idea is to first select decision pairs whose corresponding dynamics can lead to stable fixed points $(\widetilde{N}_a,\widetilde{N}_b)$; we can then select among them  those that are best in sustaining group representation.

\subsubsection{Combined Effects on Group Representation and Features}

In practice, decisions can simultaneously impact both group representation and the evolution of features, (potentially) making a bad situation worse. Consider the lending example where a lender decides whether or not to approve a loan application based on the applicant's credit score. It has been shown in \cite{pmlr-v80-liu18c} that decisions under either \texttt{EqOpt} or \texttt{DP} can potentially lead to over issuance of loans to the less qualified (disadvantaged) group. As a result, the disadvantaged group's score distribution will skew toward higher default risk. Over time, more people from this group may stop applying for loans. The increased disproportionality between the two groups will then lead the lender to actually issue more loans (relatively) to the less qualified group to satisfy \texttt{EqOpt} or \texttt{DP} fairness, leading its score distribution to skew more toward higher default risk over time.

\cite{xueru} studies the combination of these two effects on the underlying population, i.e., the effect on group representation and the effect on how features evolve. Specifically, they consider the case where feature distributions $f_{k,t}(x,y)$ are allowed to change over time, and try to understand what happens to group representation disparity $\frac{\alpha_a(t)}{\alpha_b(t)}$ when $f_{k,t}(x,y)$ are also affected by decisions. 

Let $f_{k,t}(x,y) = g^0_{k,t}f_{k,t}^0(x) + g^1_{k,t}f_{k,t}^1(x)$ be $G_k$'s feature distribution at $t$, where $g^j_{k,t} = \mathbb{P}(Y=j|Z=k)$ and $f^j_{k,t}(x) = \mathbb{P}(X=x|Y=j,Z=k)$ at $t$. Let $G_k^j$ be the subgroup of $G_k$ with label $Y = j$. Based on the facts that individuals from the same demographic group with different labels may react differently to the same decision rule, \cite{xueru} considered two scenarios of how feature distributions are reshaped by decisions: (1) $\forall k\in\{a,b\}$, $f_{k,t}^j(x) = f_k^j(x)$ remain fixed but $g_{k,t}^j$ changes over time according to $G_k^j$'s own perceived loss; and (2) $\forall k\in\{a,b\}$, $g_{k,t}^j = g_k^j$ remain fixed but for subgroup $G_k^i$ that is less favored by decision over time (experience an increased loss), its members make extra effort such that $f_{k,t}^i(x)$ skews toward the direction of lowering their losses. In both cases, \cite{xueru} shows that representation disparity can worsen over time under common fairness intervention and such exacerbation accelerates as compared to the case when feature distributions are fixed.

\subsubsection{Fairness in Reinforcement Learning Problems}

Studies in \cite{jabbari2017fairness,wen2019fairness} capture the interaction between decisions and the underlying population via a reinforcement learning framework, where the environment is described by a Markov Decision Process (MDP), defined by a tuple $(\mathcal{S},\mathcal{A},\mathbb{P},{R},\gamma)$.  $\mathcal{S}$ is the set of states representing certain properties of individuals in the system and $\mathcal{A}$ the set of actions representing available decisions. At time $t$, the decision maker by taking action $a_t\in\mathcal{A}$ in state $s_t\in \mathcal{S}$ receives a reward $r_t={R}(s_t,a_t)\in[0,1]$.  The probability of the decision maker being in state $s_{t+1}$ at time $t+1$ is given by the transition probability matrix $\mathbb{P}(s_{t+1}|a_t,s_t)$; this is what captures the impact of decisions on the underlying population.
 \cite{jabbari2017fairness} generalizes the bandits problem studied in  \cite{joseph2016fairness}\cite{joseph2018meritocratic} to the above reinforcement learning framework, by taking into account the effects of decisions on the individuals' future states and future rewards. It slightly modifies the meritocratic fairness defined in \cite{joseph2016fairness} based on long-term rewards: a decision is preferentially selected over another only if the long-term reward of the former is higher than the latter. Under such a fairness constraint, an algorithm is proposed that can achieve near-optimality within $T_0$ time steps. The impact of fairness is reflected in $T_0$: it takes more time steps to learn a near-optimal decision rule when the fairness requirement is stricter.

\cite{wen2019fairness} studied a reinforcement learning problem under group fairness (\texttt{DP}) constraint, where the state $s_t = (x_t,z_t)$ consists of both the feature $x_t$ and the sensitive attribute $z_t\in\{a,b\}$ of the individual \rev{ who is subject to the decision maker's decision at $t$}. When action $a_t$ is taken in state $s_t$, in addition to reward $R(s_t,a_t)$ received by the decision maker, the individual also receives a reward $\rho(s_t,a_t)$. 
The \texttt{DP} constraint in \cite{wen2019fairness} requires that the expected (discounted) cumulative reward of individuals from the two groups to be approximately equal. Algorithms (model-free and model-based) are developed in \cite{wen2019fairness} for learning a decision rule that is both \texttt{DP}-fair and near-optimal.

\begin{acknowledgement}

This work is supported by the NSF under grants CNS-1616575, CNS-1646019, CNS-1739517. 

\end{acknowledgement}

\section*{Appendix}
\addcontentsline{toc}{section}{Appendix}

\bibliographystyle{plain} 
\bibliography{references}

\end{document}